\documentclass[numbers]{article}

\usepackage[final]{neurips_2021}

\usepackage[utf8]{inputenc} 
\usepackage[T1]{fontenc}    
\usepackage{hyperref}       
\usepackage{url}            
\usepackage{booktabs}       
\usepackage{amsfonts}       
\usepackage{nicefrac}       
\usepackage{microtype}      
\usepackage{xcolor}         
\usepackage{graphicx}
\usepackage{xcolor} 

\newcommand\blfootnote[1]{%
  \begin{NoHyper}%
  \renewcommand\thefootnote{}\footnote{#1}%
  \addtocounter{footnote}{-1}%
  \end{NoHyper}%
}

\title{Masked Autoencoders are Scalable Learners of Cellular Morphology}

\author{%
  Oren Kraus\thanks{Equal contribution.}
  \And
  Kian Kenyon-Dean$^{*}$
  \And
  Saber Saberian
  \And
  Maryam Fallah
  \And
  Peter McLean
  \And
  Jess Leung
  \And
  Vasudev Sharma
  \And
  Ayla Khan
  \And
  Jia Balakrishnan
  \And
  Safiye Celik
  \And
  Maciej Sypetkowski
  \And
  Chi Vicky Cheng
  \And
  Kristen Morse
  \And
  Maureen Makes
  \And
  Ben Mabey
  \And
  Berton Earnshaw
}

\begin{document}

\maketitle

\begin{abstract}
Inferring biological relationships from cellular phenotypes in high-content microscopy screens provides significant opportunity and challenge in biological research. Prior results have shown that deep vision models can capture biological signal better than hand-crafted features. This work explores how self-supervised deep learning approaches scale when training larger models on larger microscopy datasets. Our results show that both CNN- and ViT-based masked autoencoders significantly outperform weakly supervised baselines. At the high-end of our scale, a ViT\mbox{-}L/8 trained on over 3.5-billion unique crops sampled from 93-million microscopy images achieves relative improvements as high as 28\% over our best weakly supervised baseline
at inferring known biological relationships curated from public databases. 
\blfootnote{All authors contributed to this article during their employment with Recursion. For correspondence, email: \url{oren.kraus@recursionpharma.com}, \url{berton.earnshaw@recursionpharma.com}, or \url{info@rxrx.ai}.}
Relevant code and select models released with this work can be found at: \url{https://github.com/recursionpharma/maes_microscopy}.
\end{abstract}

\section{Introduction}
\label{sec:intro}

A fundamental challenge in biological research is quantifying complex cellular phenotypes and relating them across genetic and chemical perturbations \cite{PrzybylaNatGeneticsReviews2022,VincentPhenotypic2022}.
Image\mbox{-}based profiling has proven to be a powerful approach for exploring cellular phenotypes induced by genetic and chemical perturbations \cite{BoutrosHCS2015}.
These experiments use \textit{high content screening} (HCS) systems combining automated microscopy with high throughput technologies to assay perturbations on a massive scale.
Recent public releases of HCS image sets, like RxRx3 \cite{fay2023rxrx3} and JUMP-CP \cite{chandrasekaran2023jump}, consist of millions of cellular images across 100,000s of unique perturbations and demonstrate the scalability of this approach. 

HCS image sets are often analyzed with customized cell segmentation, feature extraction, and downstream analysis pipelines \cite{10.1038/nmeth.4397}.
Despite the many discoveries made using this approach \cite{BoutrosHCS2015}, developing robust segmentation and feature extraction pipelines using open-source software packages \cite{10.1186/gb-2006-7-10-r100,10.1186/s12859-021-04344-9} remains challenging \cite{ChandrasekaranHCSMLupgrade2021}.
Alternatively, representation learning approaches do not require prior knowledge of cellular morphology and perform significantly better on practical biological research objectives, e.g. inferring relationships between perturbations \cite{celik2022biological}.
In contrast to previous approaches employing weakly supervised pretraining \citep{Moshkov2022}, in this work we train masked autoencoders (MAEs) \cite{he2022masked} on progressively larger HCS image sets and show that these models are scalable learners of cellular morphology, outperforming previous state-of-the-art methods at inferring known biological relationships in whole-genome HCS screens.

\section{Related Work}
\label{sec:related-work}

\textbf{Supervised learning on HCS image sets.}
Deep learning models have been successfully trained to perform cell segmentation \cite{Valen2016, Moen2019,Cellpose2021} and phenotype classification \cite{Kraus2016,Kraus2017,Ouyang2019,Eulenberg2017}, however these supervised learning tasks require the costly creation of segmentation masks and other labels.
Inspired by the successful use of embeddings obtained from ImageNet-trained models for other datasets and tasks \cite{Razavian2014}, researchers used models trained on natural images to to featurize HCS data with varying results \cite{TVN2017,Pawlowski2016}. 
Others \cite{Moshkov2022,sypetkowski2023rxrx1,saberian2022deemd} have trained convolutional networks to classify labels obtained from experimental metadata (e.g., perturbation class), a technique called \textit{weakly supervised learning} (WSL) \cite{zhou2018brief}.
Despite obtaining SOTA results when trained on small, highly-curated image sets, we show that the performance of WSL models does not necessarily improve on larger datasets.

\textbf{Self-supervised learning.}
Vision models pretrained with self-supervised learning (SSL) often outperform supervised models on downstream tasks \cite{he2022masked,caron2021dino,Chen2020simclr}. 
Unlike supervised pretraining~\cite{Kolesnikov2019bigtransfer}, SSL is readily applied to large datasets where labels are lacking or heavily biased.
This is useful for HCS datasets, as they contain a wide range of cellular phenotypes that are difficult for human experts to interpret and annotate. 
For example, DiNO~\cite{caron2021dino} is an SSL method that has been applied to HCS \cite{Cross-Zamirski2022dinohcs,Haslum2022metadataguided,Sivanandan2023posh,kim2023self,doron2023unbiased} data, however it relies on augmentations inspired by natural images, which may not be applicable to HCS image sets. 
Alternatively, masked autoencoders  \cite{he2022masked} are trained by reconstructing masked patches from unmasked patches of an image (Fig. \ref{fig:recon}). MAEs have been successfully applied to images \cite{he2022masked}, audio \cite{Huang2022audiomae}, video \cite{Feichtenhofer2022videomae} and multimodal audio-video datasets \cite{Huang2022avmae}. However, previous attempts to train MAEs on HCS datasets have had limited success \cite{xun2023microsnoop,kim2023self}, due in part to limitations in compute resources and dataset size. The present work shows that MAE training scales with both model and training set size.

\section{Methods}
\label{sec:methods}

\begin{figure}
    \centering
    \includegraphics[width=\columnwidth]{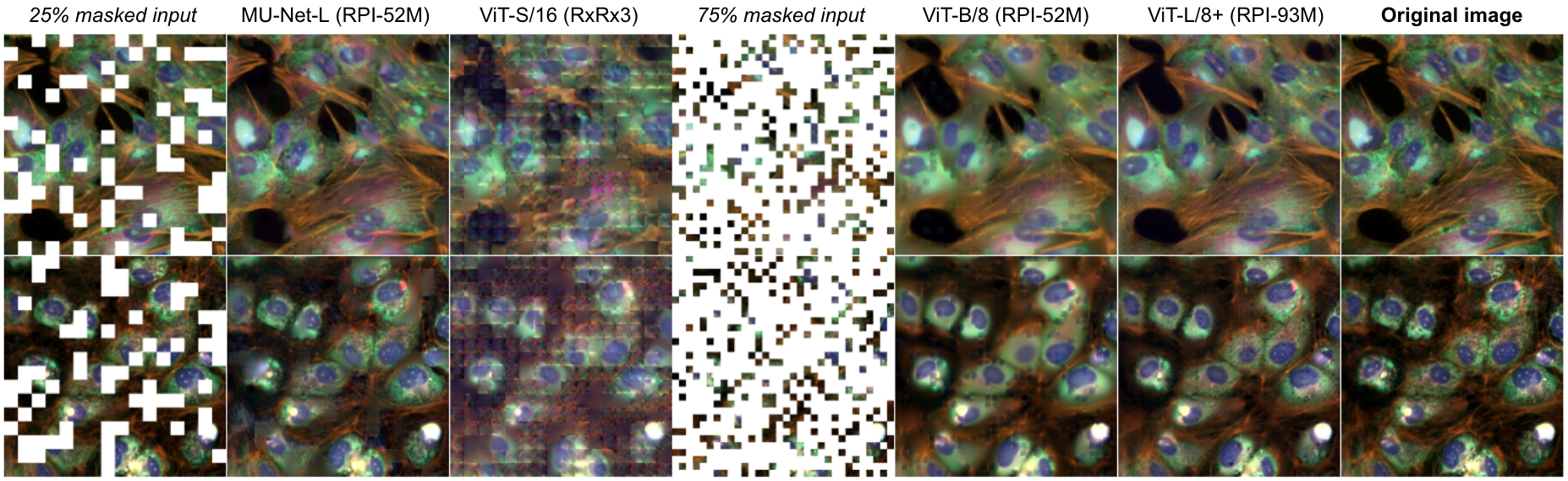}
    \caption{Visualizing reconstructions from masked random \textit{validation} images for different MAEs.}
    \label{fig:recon}
\end{figure}

\textbf{Datasets.}
We investigate the scaling properties \cite{zhai2022scaling} of cellular image sets by evaluating models trained on the following four microscopy datasets. \textbf{RxRx1} \cite{sypetkowski2023rxrx1} is a publicly-available proprietary Cell Painting dataset with 125,510 images of 4 human cell types under 1,108 different siRNA perturbations across 51 experimental batches.
\textbf{RxRx3} \cite{fay2023rxrx3} is a publicly-available proprietary Cell Painting dataset with over 2.2 million images of HUVEC cells under 17,063 CRISPR knockouts (over 6 guides) or 1,674 compounds across 180 experimental batches.
\textbf{RPI-52M} and \textbf{RPI-93M} (Recursion Phenomics Imageset) are private datasets with 52 million and 93 million proprietary Cell Painting and Brightfield images, respectively. To our knowledge, these are the largest HCS datasets collected for model training purposes. All evaluations are performed on RxRx3, which is the largest publicly available whole-genome HCS image set.

\textbf{Weakly supervised learning.} As a baseline, we employ the 28-million parameter DenseNet-161 backbone implemented in \cite{sypetkowski2023rxrx1}, trained to predict cellular perturbations and producing 128-dimensional embeddings, with and without adaptive batch normalization (AdaBN) \cite{li2018adaptive}.

\textbf{U-Nets}. 
We adapt U-Nets~\cite{ronneberger2015u} for masked autoencoding (MU-Nets) by training to reconstruct masked sections of input images. We train MU-Nets as described in \citet{xun2023microsnoop} and report results for MU-Net-M and MU-Net-L, which have 52- and 135-million parameters, respectively. MU-Net-M's downsampling scale is set to 32/64/128/256/512. MU-Net-L incorporates an additional scale of 1024. In each case, the decoder mirrors the encoder's scale configuration. After an initial hyperparameter search (see ~\ref{MUNets-appendix}), we trained both models with a mask ratio of 25\% and kernel size of 5.

\textbf{Vision transformers.}
We train vision transformers (ViTs) \cite{dosovitskiy2020image,steiner2021train,dehghani2023scaling,zhai2022scaling} as MAEs following the implementation in \citet{he2022masked}. 
We report results for ViT-S, ViT-B, and ViT-L encoders \cite{dosovitskiy2020image}, containing 22-, 86-, and 304-million parameters, respectively, and producing 384-, 768-, and 1024-dimensional embeddings, respectively. We explore the use of 8x8 and 16x16 patch sizes and 75\% and 25\% mask ratios (Fig. ~\ref{fig:recon}). A 25-million parameter decoder \cite{he2022masked} is used for patch reconstructions.
Note that 8x8 patches induce a sequence length 4 times greater than 16x16 patches and is thus more computationally expensive. 

\textbf{Training.}
Models were trained  on Recursion's HPC cluster, BioHive-1, for up to 100 epochs on as many as 128 80GB-A100 GPUs, depending on the size of the model and dataset. 256 x 256 x 6 image crops were randomly sampled from 2048 x 2048 x 6 images, augmenting with random horizontal and vertical flips. For each dataset, we use a validation set of center-cropped images from full experiments unseen during training.

\textbf{Scaling to ViT-L/8+.}
We scale training based on the results of smaller models trained on smaller datasets \cite{dehghani2023scaling,hestness2017deep,openai2023gpt4,zhai2022scaling}, as visualized in Figure~\ref{fig:scaling} (total FLOps is based on \citet{touvron2022three}). Our largest model, ViT-L/8+, was trained for over 20,000 GPU hours, learning from over 3.5 billion image crops sampled from RPI-93M. Inspired by \cite{xie2022masked}, we added a  term to the loss function to prevent divergence and improve texture reconstruction.

\textbf{Inference.}
The metrics of Section~\ref{sec:results} are calculated on the gene knockout experiments of RxRx3 \cite{fay2023rxrx3}, requiring the embedding of \textasciitilde{}140 million image crops for each encoder. See \ref{inference-appendix} for details.

\section{Results}
\label{sec:results}

\begin{figure}
    \centering
    \begin{minipage}{0.49\textwidth}
        \centering
        \includegraphics[width=0.99\textwidth]{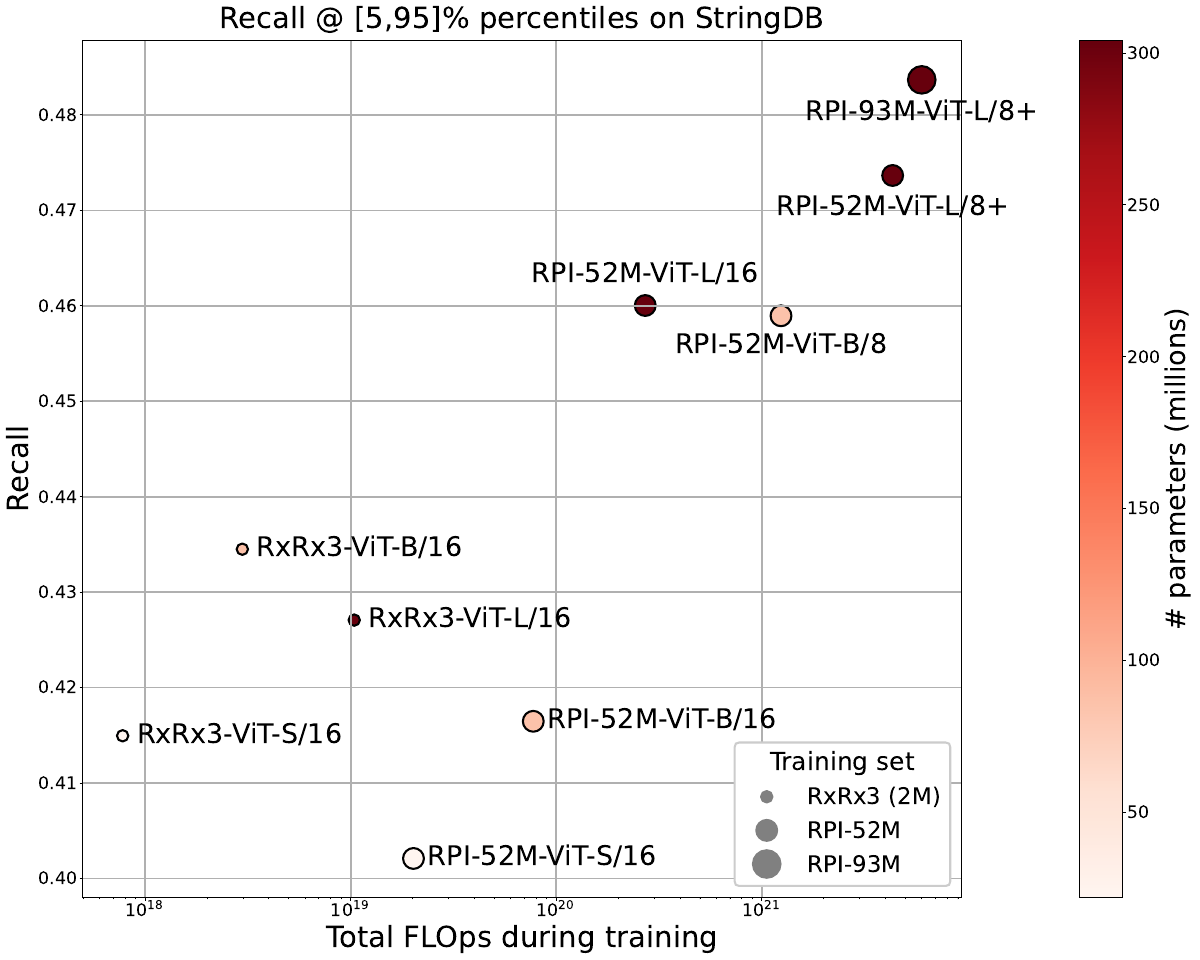}  
        \caption{StringDB recall as a function of training FLOps. Similar results hold for other databases.}
        \label{fig:scaling}
    \end{minipage}%
    \begin{minipage}{0.02\textwidth}
        \ 
    \end{minipage}%
    \begin{minipage}{0.49\textwidth}
        \centering
        \includegraphics[width=0.99\textwidth]{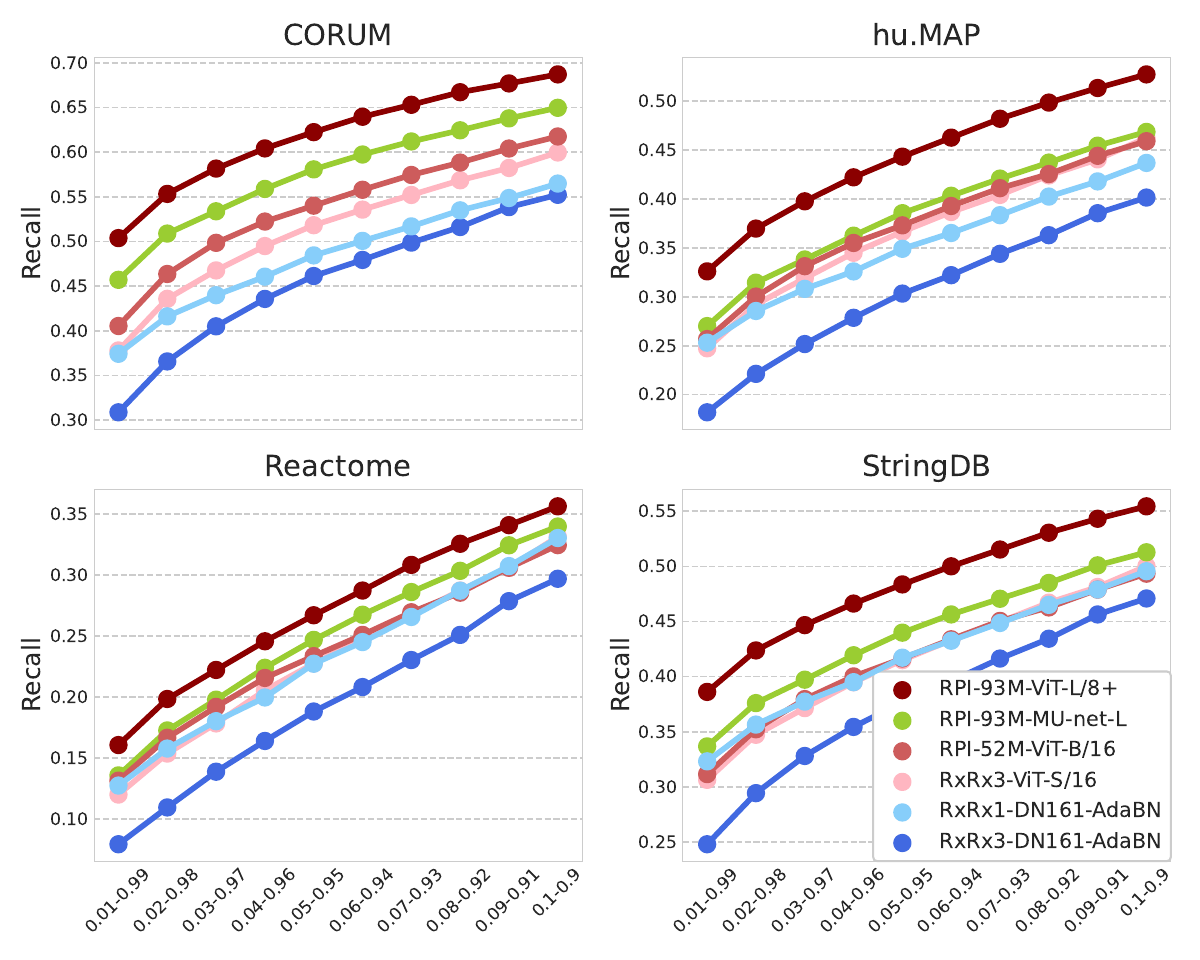} 
        \caption{Recall across different cosine similarity percentiles for each database.}
        \label{fig:benchmarks}
    \end{minipage}    
\end{figure}

An important use of HCS data is the accurate inference of biological relationships amongst genetic and chemical perturbations.
We evaluate each model's ability to capture known relationships using the multivariate metrics described in \citet{celik2022biological}. 
Briefly, each model's embeddings are first aligned across experimental batches using TVN (typical variation normalization)~\cite{TVN2017}, fitted to the negative experimental controls across all batches. 
Following TVN, we correct for possible chromosome arm biases known to exist in CRISPR-Cas9 HCS data \cite{lazar2023high}. We compute the embedding of each perturbation by taking the spherical mean over its replicate embeddings.
We use the cosine similarity of a pair of perturbation representations as a relationship metric, setting the origin of the space to the mean of negative experimental controls.
We compare these similarities with the annotated relationships found in the following public databases: CORUM~\cite{CORUM}, hu.MAP~\cite{hu_map}, Reactome~\cite{REACTOME}, and StringDB~\cite{stringdb} (with >95\% combined score).

Table~\ref{table:scaling} reports the recall of known relationships amongst the top and bottom 5\% of all cosine similarities between CRISPR knockout representations in RxRx3.
Note how both recall and image reconstruction (see Fig.~\ref{fig:recon}) improve with larger models, larger training sets, smaller patches, and larger mask ratio.
In Figure~\ref{fig:scaling} we see that recall strongly correlates with training FLOps, a function of both model and training set size (see \ref{sec:additional} for similar results trends on other databases).
Figure~\ref{fig:benchmarks} shows similar trends in recall for other similarity percentiles.
In contrast, the performance of re-implemented WSL baselines \cite{sypetkowski2023rxrx1} decreases when the dataset is scaled from RxRx1 to RxRx3, which could be due to the chromosome arm bias present in CRISPR-Cas9 systems \cite{lazar2023high} or other factors such as the increased size of the label set.

We compare these models with recent results from an alternative HCS platform combining pooled CRISPR screening with Cell Painting \cite{Sivanandan2023posh}. Table~\ref{table:poshcp-comparison} reports recall at 5\% FPR in StringDB on three gene sets defined in \citet{Sivanandan2023posh}. The ViT-L/8+ MAE trained on RPI-93M yields a minimum 20\% relative improvement in gene set performance over CP-DiNO 1640 (a ViT-S/8), which was trained on \textasciitilde{}1.5 million single-cell images. We note the significant differences in assay technology, cell lines, and modeling methodology between the two platforms, making their direct comparison impossible using this metric.
Nonetheless, we hope this comparison brings the field closer to an accepted set of benchmarks for evaluating models trained on HCS datasets.

\begin{table}[t]
\centering
\caption{Recall of known relationships in top and bottom 5\% of cosine similarities by model backbone and training set, with results for each database (CORUM/hu.MAP/Reactome/StringDB). DenseNet-161 backbones are trained via WSL, all others via SSL. See Fig.~\ref{fig:benchmarks} for recall at other percentiles.} 
\begin{tabular}{lcccc}
\toprule
\textbf{Model backbone} & RxRx1 \cite{sypetkowski2023rxrx1} & RxRx3 \cite{fay2023rxrx3} & RPI-52M & RPI-93M \\
\midrule
DenseNet-161 & .38/.31/.19/.33 & .36/.27/.17/.32 & -- & -- \\
DenseNet-161 w/ AdaBN & .48/.35/.23/.42 & .46/.30/.19/.38 & -- & -- \\
\midrule
MU-Net-M  & -- & .56/.38/.23/.42 & -- & -- \\
MU-Net-L  & -- &.57/.37/.23/.43 & .58/.39/.24/.44 & .58/.39/.25/.44 \\
MAE ViT-S/16 & -- & .52/.37/.23/.41 & .51/.36/.22/.40 & -- \\
MAE ViT-B/16 & -- & .57/.39/.23/.43 & .54/.37/.23/.42    & -- \\
MAE ViT-B/8  & -- & -- & .60/.40/.25/.46 & -- \\
MAE ViT-L/16 & -- & .56/.37/.23/.43 & .61/.41/.26/.46 & -- \\
MAE ViT-L/8+ & -- & -- & .61/.42/\textbf{.27}/.47 & \textbf{.62}/\textbf{.44}/\textbf{.27}/\textbf{.48} \\
\bottomrule
\end{tabular}
\label{table:scaling}
\end{table}

\begin{table}[t]
\centering
\caption{Recall (at 5\% false positive rate) of StringDB relationships for select models on three different gene sets defined in \citet{Sivanandan2023posh}.}
\begin{tabular}{llccc}
\toprule
\textbf{Training dataset} & \textbf{Model backbone} & PoC-124 & MoA-300 & DG-1640\\
\midrule
RxRx1 \cite{sypetkowski2023rxrx1} & WSL DenseNet-161 w/ AdaBN & .79 & .\textbf{24} &  .15 \\
RxRx3 \cite{fay2023rxrx3} & MAE ViT-S/16 & .74 & .19  & .14 \\
RPI-52M & MU-Net-L  & .79 & .20 & .15 \\
RPI-93M & MAE ViT-L/8+ & .\textbf{80} & .23 & .\textbf{17} \\
\midrule
CP-1640 \cite{Sivanandan2023posh} & DiNO ViT-S/8 & .53 & .12 & .14 \\
\bottomrule
\end{tabular}
\label{table:poshcp-comparison}
\end{table}

\section{Conclusion}
\label{sec:conclusion}

This work demonstrates that scaling properties \cite{zhai2022scaling} apply to learning representations of cellular morphology that can accurately infer known biological relationships. Unlike previous approaches that use weakly supervised learning \cite{Moshkov2022, sypetkowski2023rxrx1} on small, curated datasets, we showed that the performance of MAEs on biologically meaningful benchmarks scales to massive HCS image sets. In future work, we will continue to scale model and training set sizes even further. We will also explore new applications of this technology beyond predicting biological relationships, with the ultimate goal of creating general-purpose foundation models of cellular biology.

\subsubsection*{Acknowledgements}
This work reflects the combined efforts of many current and former Recursion employees. Special thanks to the Recursion lab team for design and execution of the HCS experiments which fueled our datasets. Additional thanks to the Recursion HPC team for their dedicated support in keeping our cluster, BioHive-1, running effectively. We would especially like to thank the following individuals for their contributions toward this work: Dominique Beaini, Jordan Christensen, Joshua Fryer, Brent Gawryluik, Imran Haque, Jason Hartford, Alex Timofeyev, and John Urbanik.





{
\small
\bibliography{bib}
}

\newpage
\appendix
\section{Appendix}
\subsection{Model hyperparameters}
RxRx1 models were trained for 100 epochs, RxRx3 models for 50 epochs, and RPI-52M / 93M models were trained for up to 50 epochs, with early stopping depending on when validation performance plateaued. All models (except those using AdaBN) use random sampling without replacement over the full dataset to create training batches. Readers are encouraged to read \cite{sypetkowski2023rxrx1} for more details on batch construction for AdaBN models.

Models were trained on Nvidia A100-80GB GPUs with data-distributed parallel (DDP) training and PyTorch 2.0. Each model was trained on 16 to 128 GPUs, depending on the size of the model and dataset.
We accelerated training speed by training with large batch sizes, as described below. 

\subsubsection{Weakly supervised learning}
Our weakly supervised CNN models were trained as standard classifiers to predict the perturbation applied in the cellular image, given a random crop as input. We use the DenseNet-161 backbone and a neck that outputs 128-dimensional embeddings for each sample, implemented exactly following the model architecture presented in \cite{sypetkowski2023rxrx1}. We trained with a crop size of 256 x 256 x 6 and batch size of 4,096. We found that training with a specially-tuned SGD optimizer yielded the best performance for these models versus other optimizers. Furthermore, we observed that issues such as chromosome arm bias \cite{lazar2023high} become significantly more pronounced for WSL classifiers trained on large datasets like RxRx3.

Adaptive batchnorm (AdaBN) is an architectural technique to enable domain adaptation \cite{li2018adaptive}.
Our AdaBN-based DenseNet-161 classifiers are implemented with Ghost BatchNorm \cite{hoffer2017train} in order to train with larger batch sizes.
Such models can perform effective domain adaptation, but they require a specialized sampler during both training and inference time to ensure that each batch is constructed from the same plate in the experiment.

\subsubsection{Masked U-Nets} \label{MUNets-appendix}
MU-Nets trained on RxRx3 used a global batch size of 4,096, while those trained on RPI-52M and RPI-93M used a global batch size of 16,384. Each was trained using the AdamW optimizer \cite{loshchilov2017decoupled} with $\beta_1=$ 0.9 and $\beta_2=$ 0.95, weight decay of 0.05, maximum learning rate 1e-3, cyclic cosine learning rate schedule, and no gradient clipping. We experimented with different mask ratios (25\%, 50\%, 75\%) and kernel sizes (3, 5). We compared the performance on the recall of biological relationships, similar to Table \ref{table:scaling}, for these values. Changing the mask ratio or kernel size did not seem to effect the performance. 

\subsubsection{Masked Autoencoder Vision Transformers}
MAE-ViTs on RxRx3 trained with a global batch size of 4,096, while those trained on RPI-52M and RPI-93M used a global batch size of 16,384. Each used the Lion optimizer \cite{chen2023symbolic} with $\beta_1=$ 0.9 and $\beta_2=$ 0.95,  weight decay of 0.05, and no gradient clipping (based on the AdamW optimizer settings from \citet{he2022masked}). We found that training dynamics and downstream performance was significantly better with large batch sizes and the Lion optimizer versus using the recommended batch size and AdamW settings presented by \citet{SSLcookbook2023}. All ViT-S and ViT-B encoders were trained with a maximum learning rate of 1e-4 and all ViT-L encoders were trained with a maximum learning rate of 3e-5 (cosine decay schedule), based on initial experiments and recommended Lion learning rate settings presented in \cite{chen2023symbolic}. All MAE-ViTs were trained with stochastic depth \cite{SSLcookbook2023}, LayerScale \cite{SSLcookbook2023}, flash attention \cite{dao2022flashattention}, parallel scaling blocks \cite{dehghani2023scaling}, QK-normalization \cite{dehghani2023scaling}, and no QK-bias \cite{dehghani2023scaling}. Stochastic depth was set to 0.1 for ViT-S and ViT-B, and 0.3 for ViT-L. All models were initialized with random weights, as initial experiments found no benefit starting from pre-trained ImageNet weights.

\subsection{Inference} \label{inference-appendix}
Inference runs on a large-scale distributed kubernetes T4 GPU cluster. Each \textit{well} in a biology experiment is imaged as a 2048 x 2048 x 6 int8 tensor. We tile over this image, obtaining 64 unique 256 x 256 x 6 crops. Each \textit{crop} is fed-forward through the encoder, and the resultant 64 embeddings are averaged to produce a final \textit{well-aggregated embedding}. Each genetics-only \textit{experiment} in RxRx3 has 9 plates, and each \textit{plate} has 1380 wells; therefore, nearly 800,000 samples need to be fed-forward through the encoder for each experiment. Given the 175 genetics-only experiments in RxRx3, this yields roughly 140 million individual samples needing to be fed-forward through each encoder (64 crops per well x 1380 wells per plate x 9 plates per experiment x 175 experiments) in order to obtain whole-genomic representations from the model. Note that the AdaBN-based weakly supervised models require careful mini-batch construction during both training and inference, whereas the rest of our models are deterministic in producing embeddings of individual samples.

\subsection{Additional results} \label{sec:additional}

\begin{figure}
    \centering
    \begin{minipage}{0.49\textwidth}
        \centering
        \includegraphics[width=0.99\textwidth]{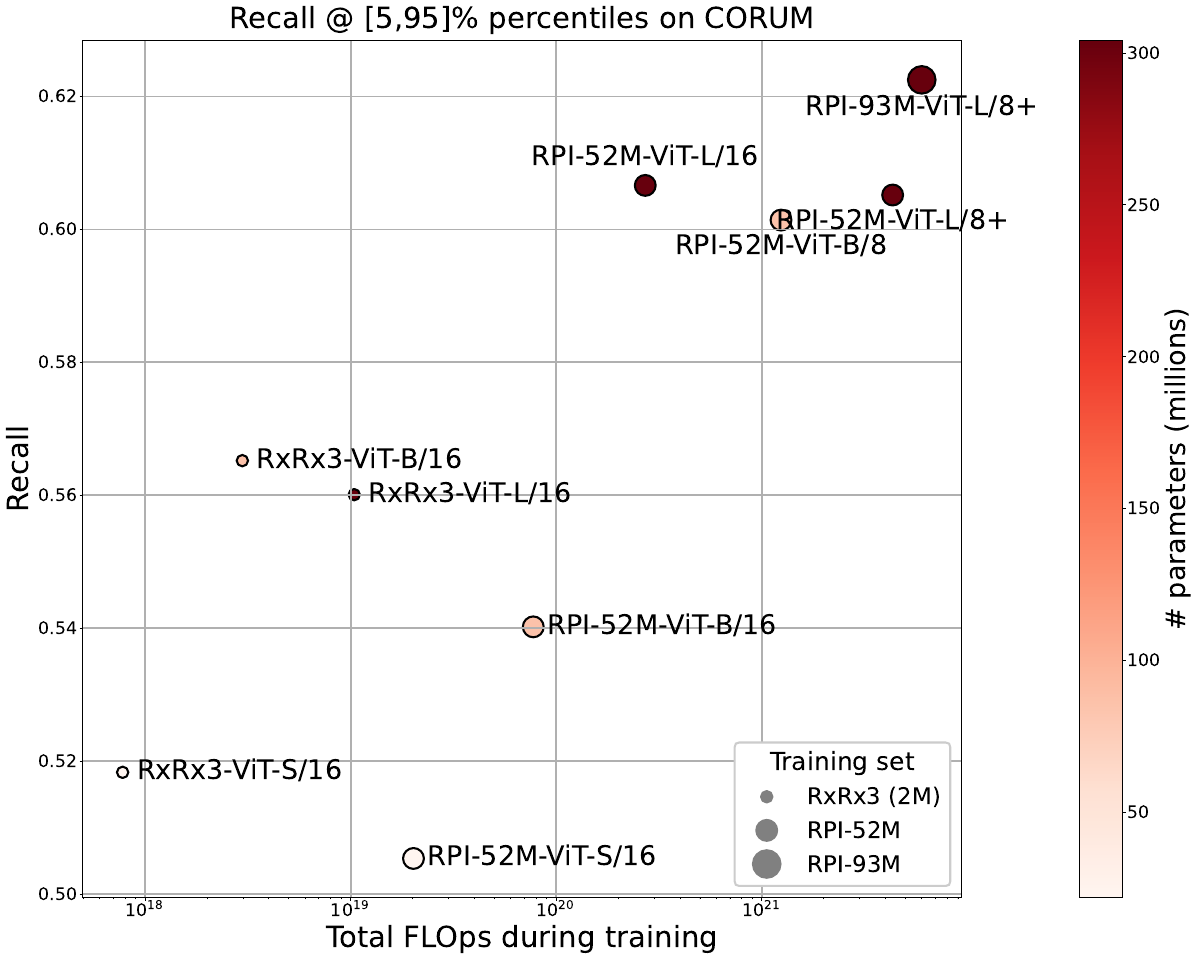}  
    \end{minipage}%
    \begin{minipage}{0.02\textwidth}
        \ 
    \end{minipage}%
    \begin{minipage}{0.49\textwidth}
        \centering
        \includegraphics[width=0.99\textwidth]{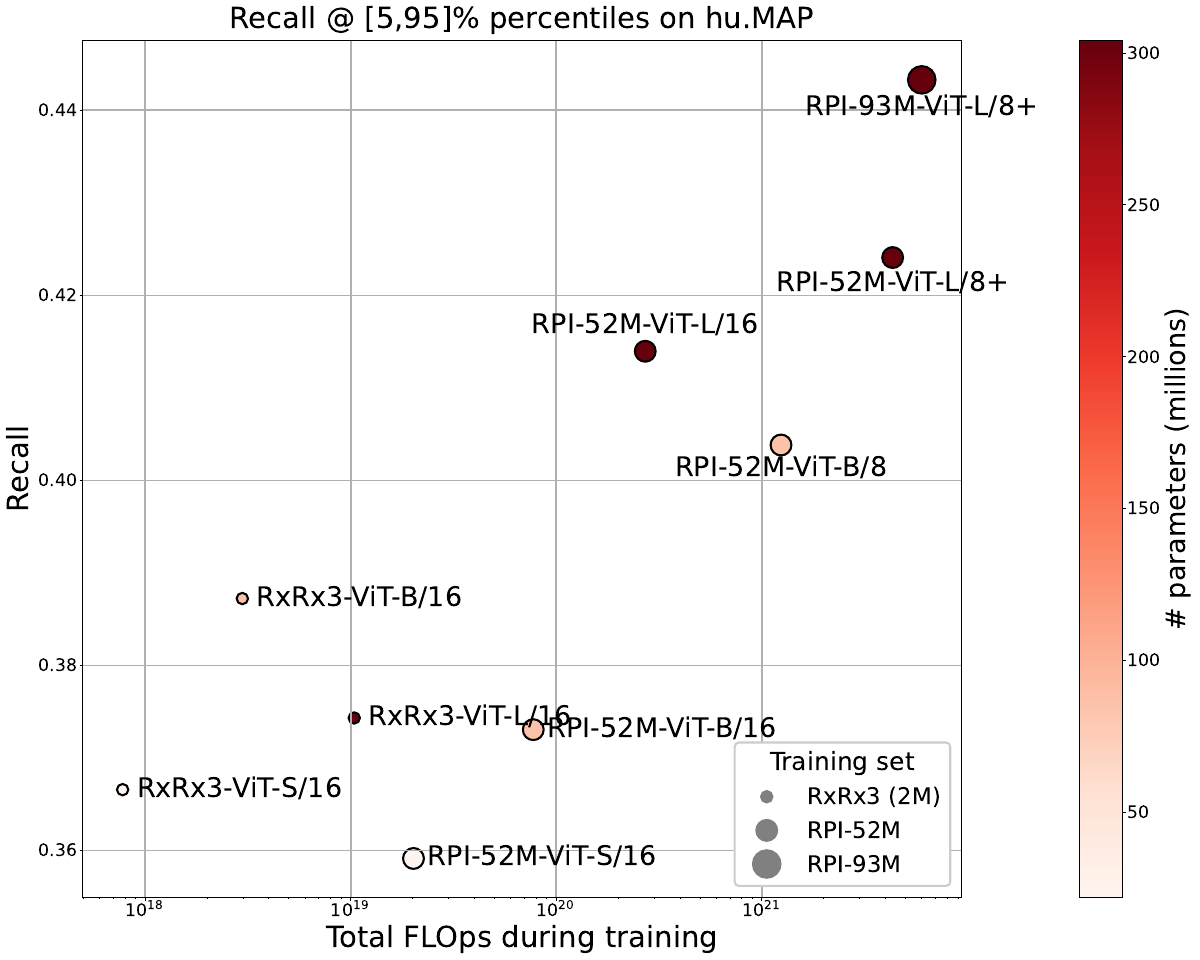} 
    \end{minipage}
    \centering
    \begin{minipage}{0.49\textwidth}
        \centering
        \includegraphics[width=0.99\textwidth]{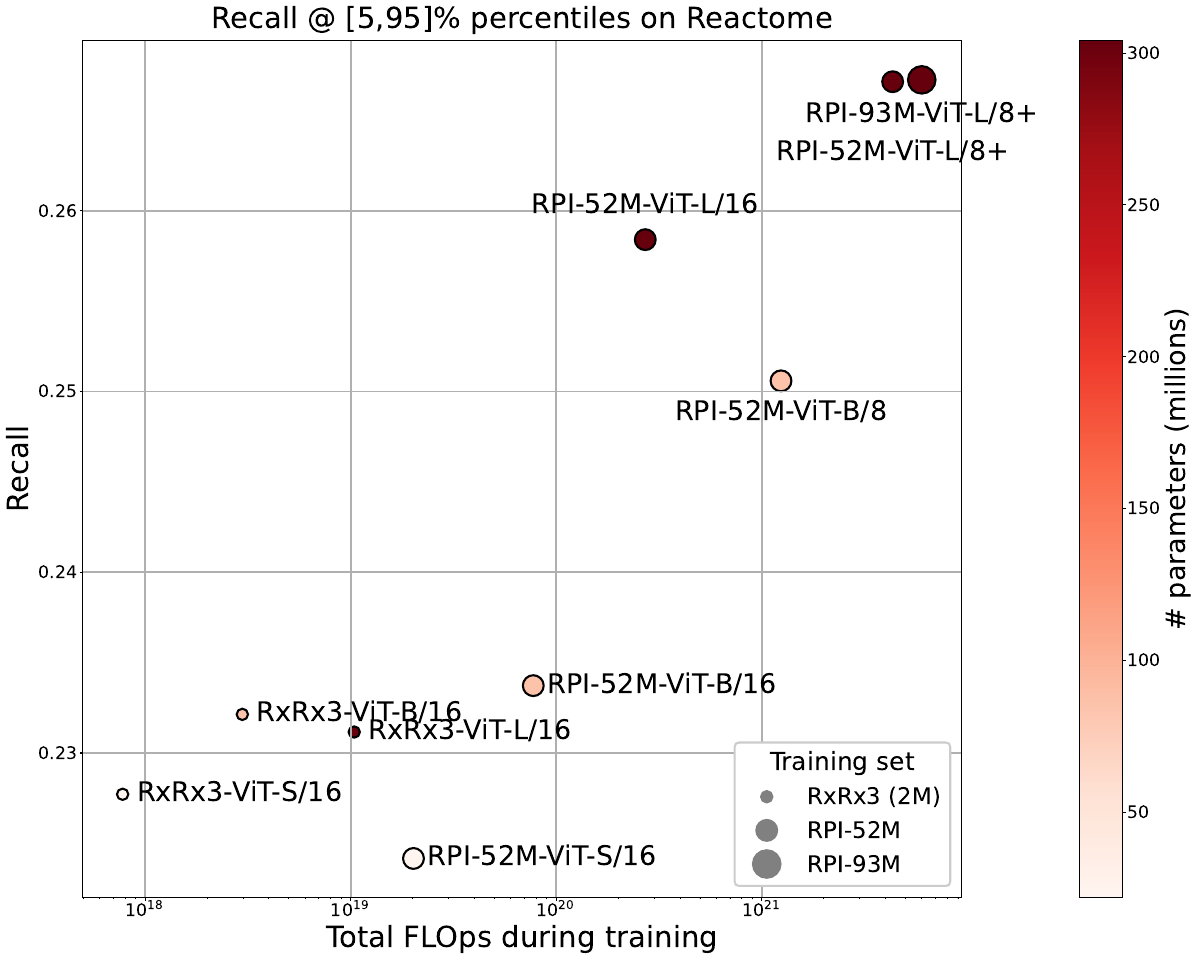}  
    \end{minipage}%
    \caption{CORUM, hu.MAP, and Reactome recalls for ViTs as a function of training FLOps.}
    \label{fig:scaling_others}
\end{figure}

\textbf{Calculation of FLOps}. In Figure~\ref{fig:scaling_others} we include the scaling plots as in Figure~\ref{fig:scaling}, for the other three benchmark databases (CORUM, hu.MAP, and Reactome). Floating point operations (FLOps) are approximated based on the FLOp counts presented in Table 1 from \citet{touvron2022three}, which presents FLOps for ViT-S/B/L/16 on a 224x224x3 image. We adjust flop counts by a factor of $(\frac{16*16}{14*14})^2 = 1.69$ to account for the changed crop size, and then for 8x8 patching models we multiply by a factor of 16 to account for the 4x more tokens and the quadratic impact this has on the attention head computations. We lastly multiply the FLOps by the number of image crops seen during training for each model.

\textbf{Recall of other baseline models}. We computed the recall metrics reported in Table \ref{table:scaling} for a number of rudimentary baselines. For example, we we generated random embeddings with different dimensions (128, 384, 512, 768, 1024), as well as randomly shuffled embeddings generated by the models illustrated in Figure \ref{fig:benchmarks}. All of these random baselines recalled  \textasciitilde{}10\% of the known relationships in each database, consistent with the fact that the metric considers only 10\% of the total cosine similarities (top and bottom 5\%). We also constructed a baseline with a simple 30-dimensional feature set built from pixel intensity statistics of each image and applied a TVN transformation to them. The recall obtained for these features, .28/.26/.16/.27 for CORUM/hu.MAP/Reactome/StringDB databases, was better than random but significantly worse than all models considered in Table \ref{table:scaling}.
\end{document}